\documentclass[letterpaper]{article} 
\usepackage{aaai2027}  
\nocopyright 
\usepackage[hyphens]{url}  
\usepackage{graphicx} 
\usepackage{natbib}  
\usepackage{caption} 
\usepackage{algorithm}
\usepackage{algorithmic}

\usepackage{newfloat}
\usepackage{listings}
\DeclareCaptionStyle{ruled}{labelfont=normalfont,labelsep=colon,strut=off} 
\floatstyle{ruled}
\newfloat{listing}{tb}{lst}{}
\floatname{listing}{Listing}

\usepackage{booktabs}

\usepackage{xspace} 
\usepackage[dvipsnames]{xcolor} 
\usepackage{amsfonts}
\usepackage{enumitem}
\usepackage{tabularx}
\usepackage{multirow}
\usepackage{colortbl}
\usepackage{pifont}
\usepackage[most]{tcolorbox}
\newtcblisting{promptbox}[2][]{
    sharp corners,
    boxrule=0.5pt,
    colback=gray!6,       
    colframe=gray!40,      
    listing only,
    listing options={
        basicstyle=\ttfamily\footnotesize, 
        breaklines=true,
        postbreak=\mbox{\textcolor{gray}{$\hookrightarrow$}\space}, 
        showstringspaces=false,
        keepspaces=true,
        columns=flexible,
    },
    title=#2,
    #1
}
\definecolor{HeaderGray}{RGB}{235, 235, 240}
\definecolor{LightBlueRow}{RGB}{248, 252, 255}
\definecolor{OursHighlight}{RGB}{245, 240, 255}
\newcommand{\method}{MetaReason\xspace}
\newcommand{\dataset}{TutorGeo\xspace}
\newcommand{\benchmark}{ExamGeo\xspace}
\newcommand{\qwenvl}{Qwen3-VL-8B-Instruct\xspace}
\newcommand{\converter}{MetaConverter\xspace}
\newcommand{\judge}{MetaJudge\xspace}
\newcommand{\reasoner}{MetaReasoner\xspace}

\title{MetaReason: Precise Interleaved Multimodal Reasoning via Editing Meta Information for Solving Geometry Problems}

\title{MetaReason: Precise Interleaved Multimodal Reasoning via Editing Meta Information for Solving Geometry Problems}
\author {
    Penghao Yin\textsuperscript{\rm 1,\rm 2},
    Haomin Wang\textsuperscript{\rm 3,\rm 2},
    Qihong Tang\textsuperscript{\rm 4},
    Xiaoye Qu\textsuperscript{\rm 2},
    Hongjie Zhang\textsuperscript{\rm 2},
    Xiao-Ping Zhang\textsuperscript{\rm 1}\corresponding
}
\affiliations {
    \textsuperscript{\rm 1}Tsinghua University\\
    \textsuperscript{\rm 2}Shanghai AI Laboratory\\
    \textsuperscript{\rm 3}Shanghai Jiao Tong University \\
    \textsuperscript{\rm 4}Nanjing University\\
    phyin2024@gmail.com
}

\begin{document}

\maketitle


\begin{abstract}
    Although visual reasoning is crucial for solving complex geometry tasks, existing vision-language models rely heavily on text-only reasoning. Some recent methods introduce intermediate visual states to facilitate reasoning, but they are often hindered by inaccurate geometric representations and low rendering fidelity, ultimately leading to unreliable outputs. To address these limitations, we propose \textbf{MetaReason}, a framework for multimodal reasoning in plane geometry that leverages structured meta-information to enable accurate auxiliary-line construction. The framework first parses geometric images into meta-information, performs controllable edits with predefined tools to synthesize high-fidelity visual states, and then conducts reasoning based on these augmented views. To support this framework, we construct \textbf{TutorGeo}, a comprehensive dataset containing 17k image-to-meta conversion samples, 60k text-only reasoning traces, and 60k interleaved multimodal reasoning traces. Using this dataset, we combine supervised fine-tuning and reinforcement learning to develop robust multimodal reasoning capabilities. We also introduce \textbf{ExamGeo}, a benchmark derived from real-world examination problems that enables systematic evaluation across varying difficulty levels. Experimental results demonstrate that MetaReason significantly outperforms existing open-source models and achieves competitive performance against proprietary models.
\end{abstract}

\begin{links}
    \link{Project}{https://github.com/PenghaoYin/MetaReason}
\end{links}

\section{Introduction}
\label{sec:intro}


Although vision language models~\cite{hurst2024gpt,chen2024internvl,bai2025qwen25vltechnicalreport,bai2025qwen3,zhu2025internvl3,wang2025internvl3} have made remarkable progress in general visual understanding, planar geometry reasoning remains a major challenge due to its reliance on fine-grained spatial perception and multi-step reasoning. Human experts often simplify such problems by constructing auxiliary lines, which make latent geometric relationships explicit and provide helpful intermediate visual structures for reasoning. In contrast, current models~\cite{gpt-5.2,claude_4_6_sonnet,gemini3_1,zhu2025internvl3,bai2025qwen3} are largely restricted to text-only reasoning, which makes it difficult to explicitly represent intermediate visual structures and often limits their performance on complex geometry tasks.

\begin{figure}[t]
  \centering
  \includegraphics[width=\columnwidth]{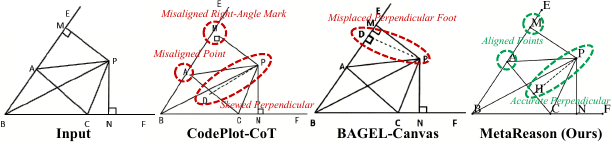}
  \caption{Comparison of auxiliary-line construction quality. Existing methods often suffer from low rendering fidelity or violate geometric constraints, while MetaReason preserves geometric relations and generate accurate diagrams.}
  \label{fig:subline_comparision}
\end{figure}

Imitating human experts, recent studies have explored interleaved multimodal reasoning by inserting images with auxiliary lines into the reasoning process~\cite{su2025thinking,su2025openthinkimg,fan2025grit,wu2025vtool,hu2024visual,duan2025codeplot,li2025zebra,shi2025mathcanvas,wang2025mathcoder}. Existing methods mainly follow two approaches, inluding code-based tool use and unified-model-based generation. Code-based methods, such as VisualSketchPad~\cite{hu2024visual} and CodePlot-CoT~\cite{duan2025codeplot}, redraw the diagram from scratch whenever auxiliary lines are needed. Since each drawing step requires a complete program, these methods introduce substantial context overhead and may produce inaccurate constructions. Unified models, such as Zebra-CoT~\cite{li2025zebra} and MathCanvas~\cite{shi2025mathcanvas}, are trained on interleaved image-text data and generate intermediate figures directly. However, current image-generation models still have difficulty preserving precise geometric relations~\cite{deng2025emerging,team2024chameleon,wang2025genexam}. As shown in Fig.~\ref{fig:subline_comparision}, both approaches can produce misaligned points and auxiliary lines. These limitations motivate an auxiliary-line construction method with concise instructions that does not rely on image-generation models.

Unlike photographs captured by cameras, geometric figures are inherently structured compositions of elements (points) and their relationships (shapes, like segments and circles). From this perspective, drawing auxiliary lines can be viewed as adding new elements or relations to an existing geometric structure. Instead of operating in pixel space or complex code space, we represent geometric figures with meta-information and construct auxiliary lines by modifying this representation. Specifically, the meta-information is a JSON-formatted structure consisting of points and shapes, with details illustrated in the Appendix. Each meta-information instance can be deterministically rendered into a geometric figure by a predefined program. To constrain the reasoning model's action space and ensure valid auxiliary-line construction, we define a toolbox that covers three common operations in geometry problem solving: \textit{drawing line segments}, \textit{constructing intersection points}, and \textit{drawing perpendicular lines}. During inference, the model can invoke these tools by emitting commands wrapped in \texttt{<tool\_call>} tags.


Following the analysis and preparatory work, we propose \method, a unified pipeline for planar geometry resolution featuring three core modules: \textbf{\converter}, \textbf{\judge}, and \textbf{\reasoner}. Addressing the intractable search space of auxiliary-line construction, we employ a two-stage training paradigm combining supervised fine-tuning and reinforcement learning. By designing a dedicated geometry reward mechanism, we guide the model to learn the optimal tool-calling policy within a large action space.


To facilitate robust training and evaluation, we introduce a large-scale planar geometry dataset, \textbf{\dataset}, alongside a challenging benchmark, \textbf{\benchmark}. \dataset comprises 137k high-quality instances, evenly split between multimodal and text-only reasoning trajectories. These trajectories are meticulously reconstructed using state-of-the-art VLMs and subsequently validated by human experts to ensure rigorous logical consistency. Furthermore, \benchmark contains 1,000 challenging problems spanning four difficulty levels defined by human expert success rates. Extensive experiments demonstrate that \method achieves 56.1\% accuracy on \benchmark, representing a 43.0\% absolute improvement over the base model and significantly outperforming both leading open-source models and several advanced proprietary models. Additionally, \method establishes new state-of-the-art performance among similarly sized models on prior benchmarks like GeoQA~\cite{chen2021geoqa} and GeoLaux~\cite{fu2025geolaux}. The main contributions of this paper are as follows:

\textbf{(1)} We propose \textbf{\method}, a visual interleaved reasoning framework based on meta-information editing, which replaces costly image or code generation with lightweight tool invocation for precise auxiliary-line construction in planar geometry.

\textbf{(2)} We introduce \textbf{\dataset}, a large-scale and high-quality training set for planar geometry reasoning, and \textbf{\benchmark}, a challenging benchmark with fine-grained difficulty levels defined by human expert success rates.

\textbf{(3)} We introduce a \textbf{two-stage training paradigm} that combines supervised fine-tuning and reinforcement, enabling \method to master tool-augmented reasoning for complex geometry problems.

\textbf{(4)} Extensive experiments demonstrate the effectiveness of \textbf{\method}, which consistently outperforms strong baselines in planar geometry reasoning.

\section{Related Works}
\label{sec:related_works}

\subsection{Datasets and Benchmarks for Multimodal Geometry Problem Solving}


With the rapid advancement of VLMs~\cite{bai2025qwen25vltechnicalreport,bai2025qwen3,zhu2025internvl3,wang2025internvl3,claude_4_6_sonnet,google2025gemini3,chen2024internvl}, a series of multimodal reasoning benchmarks have been proposed to evaluate mathematical reasoning. Early works such as Geometry3K~\cite{lu2021inter} and GeoQA~\cite{chen2021geoqa} established foundational settings for multimodal geometry understanding and reasoning. More recent benchmarks, including MMMU~\cite{yue2024mmmu}, MathVista~\cite{lu2023mathvista}, Math-Vision~\cite{wang2024measuring}, and MathVerse~\cite{zhang2024mathverse}, further broadened the evaluation scope by covering diverse mathematical problems with rich visual contexts. GeoLaux~\cite{fu2025geolaux} takes a step further toward more challenging geometry reasoning by introducing problems that require auxiliary-line construction and long-horizon deduction.

Nevertheless, existing benchmarks remain inadequate for systematically evaluating plane geometry reasoning. General multimodal math benchmarks, such as MMMU, MathVista, Math-Vision, and MathVerse, contain only a limited proportion of plane geometry problems and are not specifically designed for this setting. Meanwhile, geometry-specific benchmarks such as Geometry3K and GeoQA mostly involve relatively straightforward problems and rarely require auxiliary-line construction, limiting their ability to distinguish advanced geometric reasoning. These limitations motivate the development of a dedicated benchmark for plane geometry problem solving, particularly one centered on reasoning through auxiliary-line construction.

\begin{figure*}[t]
  \includegraphics[width=\textwidth]{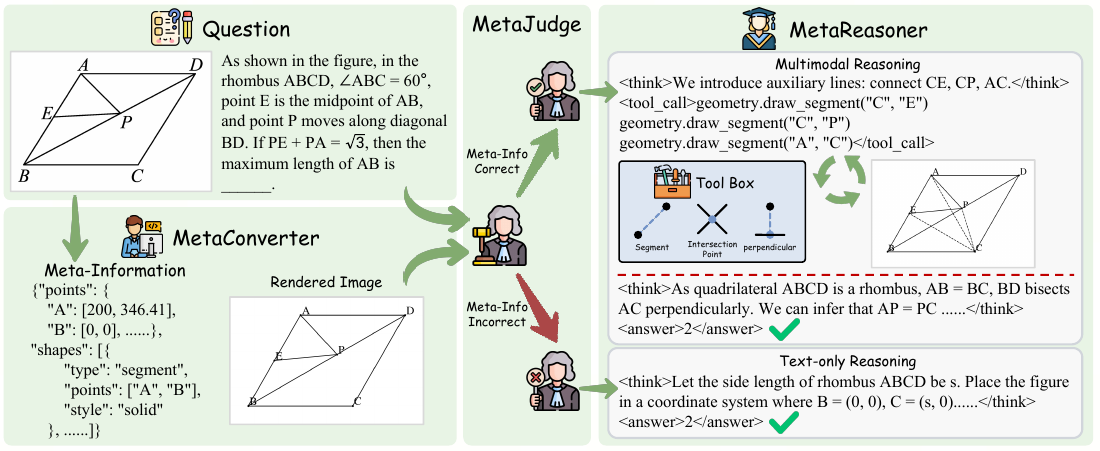}
  \caption{Overview of the MetaReason Framework. The geometric question is first converted to meta-information by MetaConverter, followed by rationality verification via MetaJudge. If reasonable, MetaReasoner implements interleaved multimodal reasoning with a pre-defined toolbox. Otherwise, it falls back to text-only reasoning.}
  \label{fig:pipeline}
\end{figure*}

\subsection{Visual Reasoning for Mathematical Geometry Problem Solving}

Some recent works introduce intermediate visual states through lightweight operations such as cropping, bounding boxes, and zooming. OpenAI o3~\cite{gpt-o3} can crop and transform images during reasoning. Chain-of-Focus~\cite{zhang2025chain} performs adaptive focusing and zooming on key regions. DeepEyes~\cite{zheng2025deepeyes} improves fine-grained understanding through active grounding and localized observation. Point-RFT~\cite{ni2025point} incorporates point-based grounding into the chain of thought. MINT-CoT~\cite{chen2025mint} further aligns visual regions with individual reasoning steps. While these methods help alleviate issues such as limited visibility and insufficient attention, they still operate by re-examining and re-annotating existing images rather than constructing new intermediate visual states. Consequently, they are fundamentally limited in tasks that require new visual structures to explicitly reveal implicit geometric relationships.

More recently, several approaches have attempted to explicitly construct auxiliary lines as intermediate visual states for reasoning. Existing methods primarily employ two approaches to construct such images: code-based tool use and unified-model-based generation. The first approach uses external tools to render auxiliary lines by generating executable plotting code, as in Visual Sketchpad~\cite{hu2024visual} and CodePlot-CoT~\cite{duan2025codeplot}. Although more expressive, these methods depend heavily on the model's coding ability and execution reliability, and they require generating complete programs at each step, resulting in substantial context overhead. Methods such as Zebra-CoT~\cite{li2025zebra} and MathCanvas~\cite{shi2025mathcanvas} rely on unified models trained on interleaved image-text reasoning data to generate intermediate figures directly. However, current image generation models still lack the precision required for geometric construction, making it difficult to faithfully translate geometric reasoning into accurate visual representations~\cite{wang2025genexam}. These limitations highlight the need for a more reliable paradigm for auxiliary-line construction.

\section{\method}
\label{sec:method}


\subsection{Overview of the Reasoning Process}
\label{sec:overview}

\method solves geometry tasks through a multi-stage pipeline, which follows an interleaved multimodal reasoning paradigm as ~\cite{wang2025pixel,zheng2025deepeyes,zhang2025thyme,hu2024visual,duan2025codeplot,shi2025mathcanvas,li2025zebra}. It is composed of 3 models, \converter, \judge, and \reasoner. As shown in Fig.~\ref{fig:pipeline}, given a multimodal query comprising a geometry diagram and a text question, the framework operates in the following three phases:

\textbf{Phase 1: Meta-Information Extraction.} For problems containing diagrams, \converter first attempts to parse the input image into a structured meta-information format (detailed in \S\ref{sec:meta_tools}). If the input contains no image, the model directly reverts to a pure-text reasoning paradigm.

\textbf{Phase 2: Meta Verification.} To ensure the accuracy of visual grounding, \judge evaluates the correctness of the extracted meta-information. This is achieved by re-rendering the meta-information as an image and verifying visual consistency. The meta-information is considered correct if the reconstructed image matches the original diagram exactly, or if the visual differences strictly adhere to the geometric constraints described in the problem text. If the meta-information passes the verification step, the system enters the visual sketching branch. Otherwise, it reverts to pure-text CoT reasoning to prevent error propagation.

\textbf{Phase 3: Iterative Visual Sketching.} Once verified, \reasoner begins an iterative interaction with the visual environment. At each step, the model analyzes the current textual and visual context and adds auxiliary lines by outputting a specific \texttt{<tool\_call>...</tool\_call>} command with actions. The environment executes this tool by modifying the meta-information, re-rendering the diagram, and returning the updated image. The newly rendered image, along with the previous text trajectory, is appended to the context for the next step. This loop continues until the model outputs the \texttt{[EOS]} token to provide the final answer, or until a predefined maximum limit of $10$ iterations is reached.

\subsection{Meta-Information Structure and Sketching Tools}
\label{sec:meta_tools}

The core of \method relies on a lightweight, symbolic representation of geometric diagrams and a concise set of tools to manipulate this representation.

\textbf{Meta-Information Structure.} We define the meta-information of a geometric diagram with a JSON-like schema composed of \texttt{points} and \texttt{shapes}. The \texttt{points} dictionary maps point names to their 2D coordinates (e.g., \texttt{"<pointName1>": [<x1>, <y1>]}). The \texttt{shapes} list under the keyword \texttt{type} contains geometric primitives. We define three basic shape types:
\textit{Segment}, \textit{Line}, and \textit{Circle}. A \textit{Segment} is a line segment connecting two endpoints, which requires a \texttt{"points"} array containing the names of the start and end points, alongside a \texttt{"style"} attribute (\texttt{"solid"} or \texttt{"dashed"}). A \textit{Line} represents an infinite line determined by two points, requiring a \texttt{"points"} array specifying the two points that it passes through, and a \texttt{"style"} attribute. Finally, a \textit{Circle} is defined by a center and a radius, requiring the name of the \texttt{"center"} point, a numerical \texttt{"radius"}, and a \texttt{"style"} attribute. Given meta-information, a geometric diagram can be illustrated with a rendering engine, which is a pre-defined Python function.

\textbf{Tools for Sketching.} To prevent excessive complexity in toolsets and to maintain a manageable prompt length, we empirically analyzed the frequency of auxiliary lines drawn in the training set, with details shown in the Appendix. The analysis reveals that connecting segments, finding intersection points, and drawing perpendicular lines are the most frequently used operations. Consequently, we equip the VLM with the following three core tools. First, the \textit{Draw Segment} tool connects two existing points to create a new line segment. Second, the \textit{Draw Intersection Point} tool calculates the intersection point between two linear elements (lines or segments) and assigns a new name to it. Third, the \textit{Draw Perpendicular} tool draws a perpendicular line from a given start point to a target linear element, and assigns a name to the corresponding foot of the perpendicular.

When the VLM invokes these tools, the framework automatically updates the JSON representation of the meta-information and invokes the rendering engine to generate the new visual sketches. The details of the meta-information schema and tools are provided in the Appendix.

\subsection{Training Strategy}
\label{sec:training}

To equip the model with both meta-information extraction and iterative visual reasoning capabilities, we adopt a two-stage training paradigm consisting of supervised fine-tuning and reinforcement learning. The detailed composition and construction of the training data are provided in \S\ref{sec:dataset}.

\textbf{Supervised Fine-Tuning.}
In the first stage, we separately fine-tune \converter and \reasoner. \converter is trained on image-to-meta pairs to learn structured geometric abstraction, while \reasoner is trained on interleaved multimodal reasoning trajectories together with pure-text reasoning data to acquire tool-use and step-by-step reasoning abilities.

\textbf{Reinforcement Learning.}
Following SFT, we further optimize \reasoner using the GRPO~\cite{shao2024deepseekmath} algorithm to improve multi-step planning and tool-use behavior in iterative visual reasoning. Specifically, we train the model exclusively on the interleaved multimodal reasoning subset of the training data. 

We utilize \qwenvl~\cite{bai2025qwen3} as the reward model to evaluate the generated trajectories. The reward function is designed to encourage both accuracy and proactive tool usage, bounded between $0$ and $1.2$. The total reward $R_{total}$ is formulated as follows:
\begin{equation}
    R_{total} = R_{correct} + \alpha \cdot \mathbb{I}_{perfect} + \beta \cdot \mathbb{I}_{tool}
\end{equation}
where $\mathbb{I}_{perfect}$ and $\mathbb{I}_{tool}$ are binary indicators for perfect completion and successful tool use, respectively. The specific reward components are defined below:

\begin{itemize}[leftmargin=*]
    \item \textbf{Correctness Reward ($R_{correct}$):} The base score for a fully correct answer is $1.0$. For single-question problems, the model receives $1.0$ for a correct answer and $0$ for an incorrect one. For multi-question problems, the reward is distributed across sub-problems. Following the weighting scheme established in MathCanvasBench~\cite{shi2025mathcanvas}, the score for the $i$-th sub-problem is calculated as:
    \begin{equation}
        score_i = \frac{1.3^i}{\sum_{k=1}^{n} 1.3^k}
    \end{equation}
    where $n$ is the total number of sub-problems. This scoring schema generally assigns higher weights to the later, more difficult sub-questions.
    
    \item \textbf{Perfect Completion Bonus ($\alpha$):} If the model correctly answers all sub-questions within a problem, it receives an additional bonus of $\alpha = 0.15$.
    
    \item \textbf{Tool-Use Bonus ($\beta$):} To encourage the model to sketch actively and reason visually, we provide a bonus of $\beta = 0.05$ if the final answer is completely correct \textit{and} the model successfully invokes the sketching tools during the reasoning trajectory.
\end{itemize}

\section{Dataset Construction and Benchmark}
\label{sec:dataset}

Existing datasets~\cite{duan2025codeplot,shi2025mathcanvas} that provide step-by-step solutions for planar geometry still have substantial limitations for training multimodal agents. First, their diagrams do not include explicit programmatic construction steps, making them unsuitable for direct use in intermediate visual sketching. Second, their reasoning trajectories are primarily designed for human learners and often omit detailed analysis, logical derivation, and reflective reasoning, which limits their effectiveness as supervision for VLM training. 

To address these limitations, we construct \dataset, a comprehensive dataset consisting of 17k image-to-meta conversion data, 60k text-only reasoning data, and 60k interleaved multimodal reasoning data. Furthermore, we introduce \benchmark, a high-quality benchmark derived from real-world examination problems.

\subsection{\dataset}
\label{sec:training_data}

\textbf{Reasoning Data Construction Pipeline.} As shown in Fig.~\ref{fig:data_process}, to transform raw geometry data into high-quality multimodal reasoning trajectories with intermediate sketches, we design a rigorous automated data processing pipeline:

\begin{figure*}[t]
  \centering
  \includegraphics[width=\linewidth]{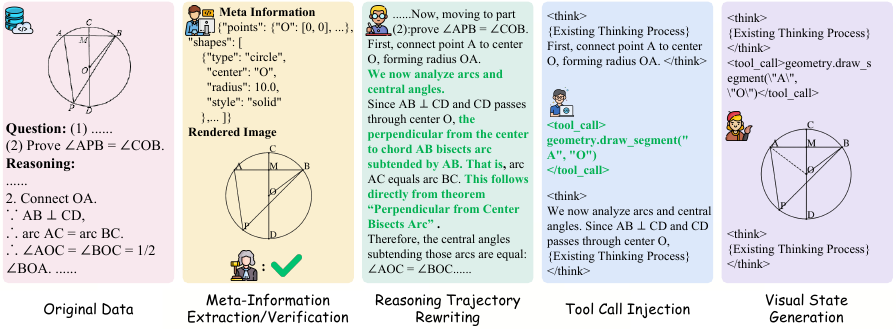}
  \caption{The \dataset Data Processing Pipeline. Raw geometric QA samples are transformed into interleaved multimodal trajectories via a four-step pipeline: (1) Meta-Information Extraction and Verification, (2) Trajectory Rewriting, (3) Tool Call Injection, and (4) Visual State Generation.}
  \label{fig:data_process}
\end{figure*}

\begin{enumerate}[leftmargin=*]
    \item \textbf{Meta-Information Extraction and Verification:} We prompt Gemini-3-Flash~\cite{google2025gemini3} to parse the original problem images into structured meta-information. Compared to open-source models, proprietary models such as Gemini-3-Flash can reliably invoke external computational tools to precisely calculate the coordinates of constructed points (e.g., intersections, midpoints, and points on angle bisectors). We then render a new diagram from the extracted meta-information and use it in place of the original image. To ensure high visual fidelity, we use \qwenvl~\cite{bai2025qwen3} as a verifier to filter out parsed meta-information that produces invalid or misaligned diagrams.
    \item \textbf{Reasoning Trajectory Rewriting:} We utilize \qwenvl to rewrite the original human-centric solutions into detailed, step-by-step reasoning processes that explicitly articulate the logical deductions required by VLMs.
    \item \textbf{Tool Call Injection:} \qwenvl is used to identify the reasoning steps at which auxiliary lines should be introduced. At these exact locations, we inject the corresponding \texttt{<tool\_call>} commands (as defined in \S\ref{sec:meta_tools}).
    \item \textbf{Visual State Generation:} Based on the injected tool call and the current meta-information, we render the updated intermediate diagram and insert it into the trajectory, forming a complete interleaved multimodal reasoning sequence.
\end{enumerate}

To ensure the model retains the ability to perform text-only reasoning when visual sketching is unnecessary or fails, we also include a subset of data that bypasses the visual generation steps and only undergoes Step 2. 


\textbf{Image-to-Meta Data.} To endow VLM with the ability to natively parse images into structured meta-information without relying on proprietary APIs during inference, we collect the detailed reasoning traces generated by Gemini-3-Flash during Step 1 of the pipeline. We use this data to train \qwenvl specifically for the \textit{image-to-meta} task.

\textbf{RL Data Selection.} During the RL stage, the training objective is strictly focused on optimizing the model’s multimodal reasoning and planning capabilities, with the goal of learning when and how to effectively construct auxiliary lines for successful problem solving. Therefore, RL is conducted on 6K interleaved multimodal reasoning samples from \dataset.

\subsection{\benchmark}
\label{sec:benchmark}

To rigorously evaluate model performance on planar geometry reasoning, we construct \benchmark, which consists of 1,000 held-out questions selected from our data collection. 

To ensure that the benchmark provides sufficient discriminative power for advanced VLMs, we categorize the questions into four difficulty levels (L1, L2, L3, and L4) based on publicly available human accuracy rates ($0.8$, $0.6$, $0.4$, and $0.2$, respectively). A lower human accuracy rate indicates a higher difficulty level. We intentionally oversample the more difficult questions to rigorously challenge the models. However, because the most difficult questions (with an accuracy rate of $0.2$) are naturally scarce in the source data, we sample the final 1,000 questions using a ratio of $1 : 2 : 5 : 2$, which corresponds to the human accuracy rates of $0.8$, $0.6$, $0.4$, and $0.2$. This curated distribution ensures that \benchmark is heavily weighted towards complex reasoning tasks, effectively distinguishing models that can genuinely ``think-with-image'' from those that rely on superficial pattern matching.

Given the structural complexity and diverse expressive formats of answers, rule-based parsing is inadequate for precise evaluation of \benchmark. Therefore, similwar to ~\cite{shi2025mathcanvas}, we employ an LLM-as-a-judge framework, utilizing GPT-5.2 as the automated evaluator to assess final answers. Detailed evaluation prompts are provided in the Appendix.

\section{Experiments}
\label{sec:experiments}

\begin{table*}[t]
\centering
\caption{Comparison of model performance on \benchmark, GeoLaux-mini~\cite{fu2025geolaux} and GeoQA~\cite{chen2021geoqa} benchmarks. The abbreviation MM denotes the multimodal reasoning paradigm. \benchmark consists of four difficulty levels from ``L1'' to ``L4'', with higher numbers indicating greater difficulty. The best and second-best results are indicated in \textit{bold} and \underline{underlined} text, respectively. All results are reported as percentages (\%).}
\label{tab:main_results}
\resizebox{\textwidth}{!}{
\begin{tabular}{l|c|c|ccccc|c|ccccc}
\toprule
\multirow{2}{*}{\textbf{Models}} & \multirow{2}{*}{\textbf{\#Params.}} & \multirow{2}{*}{\textbf{MM}} & \multicolumn{5}{c}{\textbf{ExamGeo}} & \textbf{GeoLaux-mini} & \multicolumn{5}{c}{\textbf{GeoQA}} \\
\cmidrule(lr){4-8} \cmidrule(lr){9-9} \cmidrule(lr){10-14}
 & & & L1 & L2 & L3 & L4 & \textbf{Avg.} & \textbf{Accuracy} & Angle & Area & Length & Other & \textbf{Avg.} \\
\midrule
\multicolumn{14}{c}{\textit{Closed-source Models}} \\
\midrule
\rowcolor{LightBlueRow} GPT-5.2~\cite{gpt-5.2}             & -   & \textcolor{Red}{\ding{55}}  & 67.0 & \underline{61.5} & 44.8 & 36.5 & \underline{48.7} & \underline{76.9} & 88.3 & 88.9 & 93.8 & \underline{88.4} & 90.4 \\
Claude-Sonnet-4.6~\cite{claude_4_6_sonnet}   & -   & \textcolor{Red}{\ding{55}}  & 64.0 & 55.0 & \underline{45.0} & \underline{40.5} & 48.0 & 56.1 & \underline{93.8} & \underline{91.0} & \underline{97.0} & 85.5 & \underline{94.7} \\
\rowcolor{LightBlueRow} Gemini-3.1-Pro~\cite{gemini3_1}      & -   & \textcolor{Red}{\ding{55}}  & \textbf{80.0} & 53.5 & 28.0 & 16.5 & 36.0 & 67.9 & \textbf{96.7} & \textbf{95.1} & \textbf{98.3} & \textbf{91.3} & \textbf{97.1} \\
\midrule
\multicolumn{14}{c}{\textit{Open-source Models}} \\
\midrule
InternVL3.5-8B~\cite{wang2025internvl3}      & 8B  & \textcolor{Red}{\ding{55}}  & 27.0 & 25.5 & 15.8 & 11.0 & 17.9 & 38.9 & 66.5 & 59.4 & 70.4 & 69.6 & 67.6 \\
\rowcolor{LightBlueRow} Qwen3-VL-8B-Instruct~\cite{bai2025qwen3} & 8B  & \textcolor{Red}{\ding{55}}  & 12.0 & 14.0 & 13.4 & 12.0 & 13.1 & 28.5 & 57.3 & 64.7 & 69.2 & 79.7 & 62.5 \\
\rowcolor{LightBlueRow} Bagel~\cite{deng2025emerging}              & 7B  & \textcolor{Green}{\ding{51}} & 53.0 & 35.0 & 15.6 & 13.0 & 22.7 & 44.8 & 65.1 & 66.9 & 64.2 & 50.7 & 64.7 \\
Bagel-Zebra-CoT~\cite{li2025zebra}     & 7B  & \textcolor{Green}{\ding{51}} & 51.0 & 28.0 & 10.1 & 8.0  & 17.3 & 37.6 & 52.2 & 57.0 & 65.0 & 43.5 & 57.2 \\
\rowcolor{LightBlueRow} BAGEL-Canvas~\cite{shi2025mathcanvas}          & 7B  & \textcolor{Green}{\ding{51}} & 53.0 & 28.5 & 14.6 & 7.5  & 19.8 & 37.1 & 83.3 & 79.0 & 82.2 & 65.2 & 82.4 \\
CodePlot-CoT~\cite{duan2025codeplot}        & 32B & \textcolor{Green}{\ding{51}} & 22.0 & 17.0 & 7.8  & 6.0  & 10.7 & 14.5 & 14.4 & 14.2 & 19.5 & 18.8 & 16.4 \\
\midrule
\rowcolor{OursHighlight} \textbf{\method-SFT}  & 8B  & \textcolor{Green}{\ding{51}} & 67.0 & 49.5 & 29.2 & 26.5 & 36.5  & 62.4 & 81.2 & 69.4 & 80.0 & 63.8 & 79.8 \\
\rowcolor{OursHighlight} \textbf{\method-RL}   & 8B  & \textcolor{Green}{\ding{51}} & \underline{72.0} & \textbf{70.0} & \textbf{51.6} & \textbf{45.5} & \textbf{56.1} & \textbf{80.1} & 88.8 & 79.6 & 87.6 & 68.1 & 87.5 \\
\rowcolor{OursHighlight} $\Delta$ Over Base Model &	& &		\textcolor{Green}{+60.0} &	\textcolor{Green}{+56.0} &	\textcolor{Green}{+38.2} & \textcolor{Green}{+33.5} &	\textcolor{Green}{+43.0}  &	\textcolor{Green}{+51.6} & \textcolor{Green}{+31.5} &	\textcolor{Green}{+14.9} &	\textcolor{Green}{+18.3} &	\textcolor{Green}{-11.6} &	\textcolor{Green}{+24.9} \\
\bottomrule
\end{tabular}
}
\end{table*}

\subsection{Implementation Details}
\label{sec:exp_setup}

Both \converter and \reasoner are trained based on \qwenvl. We freeze the ViT modules~\cite{dosovitskiyimage} and fine-tune the LLM backbones via LLaMA-Factory~\cite{zheng2024llamafactory}. During SFT, \converter is trained on 17k image-to-meta pairs for 3 epochs (lr=$3 \times 10^{-5}$). \reasoner is trained on 120k high-quality trajectories (60k multimodal, 60k text-only) for 2 epochs (lr=$1 \times 10^{-5}$). In the subsequent RL phase, \reasoner is optimized on the interleaved multimodal reasoning subset utilizing GRPO~\cite{shao2024deepseekmath} via verl~\cite{sheng2024hybridflow} for 2 epochs. We evaluate \method-SFT and \method-RL, setting the inference temperature to 0.0 and allowing a maximum of 10 visual sketching iterations per problem.

\subsection{Main Results}

\textbf{Performance on \benchmark.} 
As illustrated in Tab.~\ref{tab:main_results}, \textbf{\method-RL} achieves state-of-the-art performance with an average accuracy of \textbf{56.1\%} on \benchmark, significantly outperforming all open-source and closed-source baselines. 

Notably, the proposed 8B model surpasses the most powerful proprietary model, GPT-5.2 (48.7\%), by a substantial absolute margin of 7.4\%. When comparing across difficulty levels, proprietary models like Gemini-3.1-Pro perform well on L1 questions (80.0\%) but suffer severe performance degradation on L3 (28.0\%) and L4 (16.5\%) questions. In contrast, \method-RL exhibits remarkable robustness, achieving \textbf{51.6\%} and \textbf{45.5\%} on the L3 and L4 levels, respectively. This demonstrates that while text-only reasoning is sufficient for simple problems, the explicit ``think-with-image'' paradigm enabled by the precise meta-information manipulation is crucial for solving highly complex, multi-step geometric tasks.

Compared with other open-source models employing multimodal reasoning (e.g., BAGEL-Canvas, CodePlot-CoT), the proposed method demonstrates significant superiority. Existing methods struggle heavily on \benchmark (all scoring below 25\% on average), as their low rendering fidelity and inaccurate geometric representations lead to error propagation. Furthermore, the substantial improvement from the base \qwenvl (13.1\%) to \method-SFT (36.5\%), and finally to \method-RL (56.1\%), highlights the effectiveness of both the high-quality \dataset dataset and the RL training stage in activating complex tool-use planning.

\textbf{Performance on GeoLaux-mini and GeoQA.}
To evaluate the generalization ability of \method, we further evaluate it on GeoLaux-mini and GeoQA, with results summarized in Tab.~\ref{tab:main_results}. On \textbf{GeoLaux-mini}, which explicitly tests auxiliary-line construction, \method-RL achieves state-of-the-art accuracy of \textbf{80.1\%}, surpassing GPT-5.2 (76.9\%) and significantly outperforming the best open-source competitor (Bagel at 44.8\%). This directly validates the core motivation: the pre-defined sketching tools and deterministic rendering engine effectively equip the VLM with a reliable ``ruler and compass,'' which allows the model to discover implicit geometric relationships that are often overlooked by text-only reasoning.


On \textbf{GeoQA}, \method-RL achieves an average accuracy of \textbf{87.5\%}, improving substantially over the base model (62.5\%) and outperforming the strongest open-source baseline, BAGEL-Canvas (82.4\%). Gemini-3.1-Pro achieves a higher accuracy of 97.1\%, but \method-RL remains competitive as an 8B open-source model. At the category level, \method-RL improves accuracy on Angle, Area, and Length, which together account for 98.6\% of the benchmark. Detailed category results are provided in the Appendix.

\textbf{Reliability of \converter and \judge.}
We further evaluate both modules through human annotation. \converter achieves conversion accuracy ranging from 52.5\% to 82.0\% across the three benchmarks, while \judge achieves both acceptance precision and inconsistent rejection over 90\%. Detailed results are provided in the Appendix.

\begin{figure*}[t]
  \centering
  \includegraphics[width=0.99\linewidth]{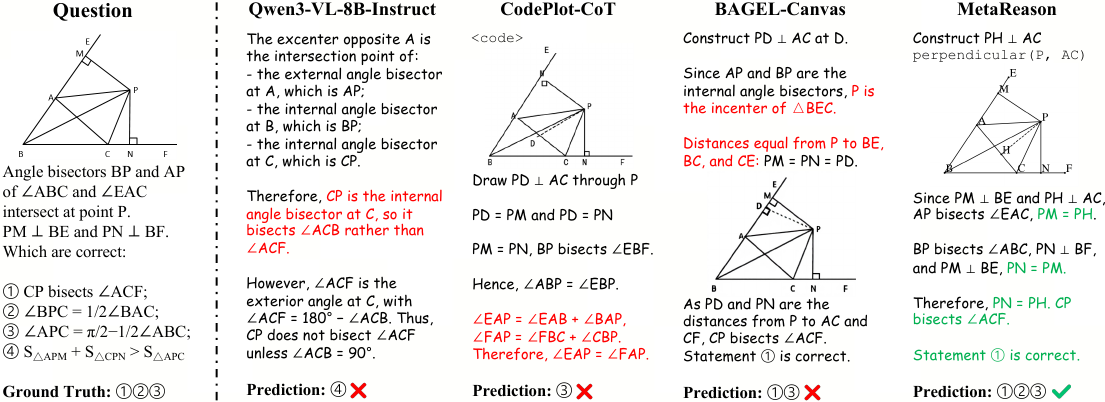}
  \caption{Case Study of Geometric Reasoning. Compared to methods that exhibit hallucinations due to the absence of auxiliary lines (Qwen3-VL-8B-Instruct) or inaccurate visual support (CodePlot-CoT and BAGEL-Canvas), our MetaReason framework generates deterministic visual states via tool-augmented meta-editing and derives the correct conclusion. For clarity, only the reasoning for the first statement is shown; the reported predictions cover all four statements.}
  \label{fig:case_study}
\end{figure*}

\begin{table*}[t]
\centering
\caption{Ablation study of the two-stage training pipeline utilizing \dataset and the \method framework across \benchmark, GeoLaux\_mini~\cite{fu2025geolaux}, and GeoQA~\cite{chen2021geoqa}. The abbreviation MM denotes the multimodal reasoning paradigm, while Training represents the two-stage training pipeline incorporating the proposed \dataset. The best and second-best results are highlighted in \textit{bold} and \underline{underlined} text, respectively. All results are reported as percentages (\%).}
\label{tab:ablation_results}
\resizebox{\textwidth}{!}{
\begin{tabular}{lcc|ccccc|c|ccccc}
\toprule
\multirow{2}{*}{\textbf{Models}} & \multirow{2}{*}{\textbf{Training}} & \multirow{2}{*}{\textbf{MM}} & \multicolumn{5}{c}{\textbf{ExamGeo}} & \textbf{GeoLaux\_mini} & \multicolumn{5}{c}{\textbf{GeoQA}} \\
\cmidrule(lr){4-8} \cmidrule(lr){9-9} \cmidrule(lr){10-14}
 & & & L1 & L2 & L3 & L4 & \textbf{Avg.} & \textbf{Accuracy} & Angle & Area & Length & Other & \textbf{Avg.} \\
\midrule
\rowcolor{LightBlueRow} \textbf{\method-RL}     &  \textcolor{Green}{\ding{52}} & \textcolor{Green}{\ding{52}} & \underline{72.0} & \textbf{70.0} & \textbf{51.6} & \textbf{45.5} & \textbf{56.1} & \textbf{80.1} & \textbf{88.8} & \textbf{79.6} & \textbf{87.6} & 68.1 & \textbf{87.5} \\
w/o \method            & \textcolor{Green}{\ding{52}} & \textcolor{Red}{\ding{55}} &  \textbf{73.0} & \underline{59.5} & \underline{44.6} & \underline{33.0} & \underline{48.1} & \underline{75.6} & \underline{78.8} & 64.4 & 77.1 & 68.1 & \underline{77.1} \\
\rowcolor{LightBlueRow} w/o \dataset training          &  \textcolor{Red}{\ding{55}} & \textcolor{Green}{\ding{52}} & 37.0 & 27.5 & 22.2 & 19.0 & 24.1 & 46.6 & 68.4 & \underline{72.1} & \underline{81.1} & \textbf{81.2} & 73.6 \\
\textcolor{Gray}{Qwen3VL-8B-Instruct}    & \textcolor{Red}{\ding{55}} & \textcolor{Red}{\ding{55}} & \textcolor{Gray}{12.0} & \textcolor{Gray}{14.0} & \textcolor{Gray}{13.4} & \textcolor{Gray}{12.0} & \textcolor{Gray}{13.1} & \textcolor{Gray}{28.5} & \textcolor{Gray}{57.3} & \textcolor{Gray}{64.7} & \textcolor{Gray}{69.3} & \textcolor{Gray}{\underline{79.7}} & \textcolor{Gray}{62.5} \\
\bottomrule
\end{tabular}
}
\end{table*}

\subsection{Qualitative Analysis}
\label{sec:case_study}

To demonstrate the superiority of the \method framework, a qualitative comparison is presented in Fig.~\ref{fig:case_study} using a geometry problem that requires auxiliary lines.



\textbf{Limitations of Baselines. }
Without visual grounding, the text-only \qwenvl model hallucinates an angle-bisector relation and fails at an early stage. CodePlot-CoT and BAGEL-Canvas attempt to construct auxiliary lines through code or image generation, but their distorted or misaligned visual states propagate errors into the subsequent reasoning.

\textbf{Success of \method.} 
In contrast, \method edits structured meta-information and deterministically renders the required auxiliary line ($PH \perp AC$) without the distortions observed in the baseline outputs. This accurate visual feedback enables the model to apply the relevant geometric relations step by step and derive the correct answer.

\subsection{Ablation Study}
\label{sec:ablation}

We ablate the contributions of the iterative visual reasoning framework and \dataset training, with results summarized in Tab.~\ref{tab:ablation_results}.

\textbf{Effect of the \method Framework.} 
The \textbf{w/o \method} variant receives the same two-stage training as \method-RL but uses pure-text CoT at inference. Removing visual reasoning reduces average accuracy by 10.4 points on GeoQA and by 4.5 points on GeoLaux-mini. On \benchmark, the two variants perform similarly on L1, whereas \method-RL gains 7.0 and 12.5 points on L3 and L4, respectively. It shows that intermediate visual construction becomes increasingly important as problem difficulty grows.

\textbf{Effect of Interleaved Multimodal Reasoning.}
Without \dataset training, the zero-shot multimodal variant still consistently outperforms the pure-text base model on all three benchmarks. This indicates that part of the gain comes from introducing intermediate visual constructions rather than from training alone.

\textbf{Effect of \dataset Training.} 
Removing \dataset training reduces accuracy from 56.1\% to 24.1\% on \benchmark and from 80.1\% to 46.6\% on GeoLaux-mini. These gaps show that training is essential for reliable meta-information generation, multi-step planning, and tool invocation.

\section{Conclusion}
\label{sec:conclusion}

In this paper, we present MetaReason, an interleaved multimodal reasoning framework that constructs auxiliary lines by editing structured meta-information with lightweight tool calls. We also introduce \dataset, a 137k-sample training set, and \benchmark, a benchmark with difficulty levels based on human accuracy. On this basis, MetaReason is trained with SFT and RL to learn multi-step planning and tool use. Experiments show that it achieves state-of-the-art performance on \benchmark and GeoLaux-mini and outperforms all open-source models on GeoQA.

\bibliography{aaai2027}

\onecolumn
\appendix
\section{Definition of Meta-Information}
The meta-information for each image consists of two primary components: \textit{points} and \textit{shapes}. \textit{points} are defined by unique names and their corresponding spatial coordinates. \textit{shapes}, including \textit{segments}, \textit{lines}, and \textit{circles}, are constructed based on \textit{points}. Detailed descriptions, definitions, and examples for each element of meta-information are provided in Table~\ref{tab:meta_info}.

\begin{figure}[!htbp]
    \centering
    \includegraphics[width=0.75\linewidth]{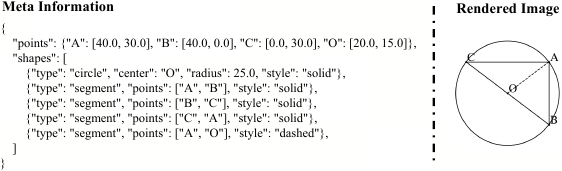}
    \caption{An Example of Meta-Information and the Corresponding Rendered Image.}
    \label{fig:meta_info}
\end{figure}

To further illustrate meta-information, Fig.~\ref{fig:meta_info} presents a concrete example where points and shapes are rendered into a geometric figure. This visualization demonstrates how our meta-information effectively captures the underlying structural logic of geometric images, ensuring a precise mapping between symbolic representations and visual layouts.

\renewcommand{\tabularxcolumn}[1]{m{#1}}
\newcolumntype{L}{>{\raggedright\arraybackslash}X}
\newcolumntype{M}[1]{>{\raggedright\arraybackslash}m{#1}}

\begin{table}[!htbp]
\centering

\caption{Definition of meta-information. This table provides detailed
specifications for geometric elements, including \textit{point},
\textit{segment}, \textit{line}, and \textit{circle}. Each element is
defined by a set of predefined fields, such as \texttt{coordinates},
\texttt{points}, and \texttt{radius}.}
\label{tab:meta_info}

\small
\renewcommand{\arraystretch}{1.2}

\begin{tabularx}{\textwidth}{
    M{1.2cm}
    L
    M{5.5cm}
    M{4cm}
}
\toprule

\textbf{Element}
&
\textbf{Description}
&
\textbf{Required Fields}
&
\textbf{Example}
\\

\midrule
\addlinespace[2pt]

\textbf{point}
&
A named point with coordinates, used to construct geometric objects such
as segments, lines, and circles.
&
\textbullet\ \texttt{name: [coord\_x, coord\_y]}
&
\texttt{"A": [10.5, 20.0]}
\\

\addlinespace[2pt]

\textbf{segment}
&
A line segment connecting two named points in the point set.
&
\textbullet\ \texttt{"type": "segment"} \par
\textbullet\ \texttt{"points": [p1, p2]} \par
\textbullet\ \texttt{"style": "solid" | "dashed"}
&
\texttt{"type": "segment",} \par
\texttt{"points": ["A", "B"],} \par
\texttt{"style": "solid"}
\\

\addlinespace[2pt]

\textbf{line}
&
An infinite line determined by two named points in the point set.
&
\textbullet\ \texttt{"type": "line"} \par
\textbullet\ \texttt{"points": [p1, p2]} \par
\textbullet\ \texttt{"style": "solid" | "dashed"}
&
\texttt{"type": "line",} \par
\texttt{"points": ["A", "C"],} \par
\texttt{"style": "dashed"}
\\

\addlinespace[2pt]

\textbf{circle}
&
A circle defined by a named center point and a radius value.
&
\textbullet\ \texttt{"type": "circle"} \par
\textbullet\ \texttt{"center": pName} \par
\textbullet\ \texttt{"radius": number} \par
\textbullet\ \texttt{"style": "solid" | "dashed"}
&
\texttt{"type": "circle",} \par
\texttt{"center": "A",} \par
\texttt{"radius": 5.0,} \par
\texttt{"style": "solid"}
\\

\addlinespace[2pt]
\bottomrule
\end{tabularx}

\end{table}

\section{Tool Selection}

\begin{figure}[H]
\centering
\includegraphics[width=0.57\linewidth]{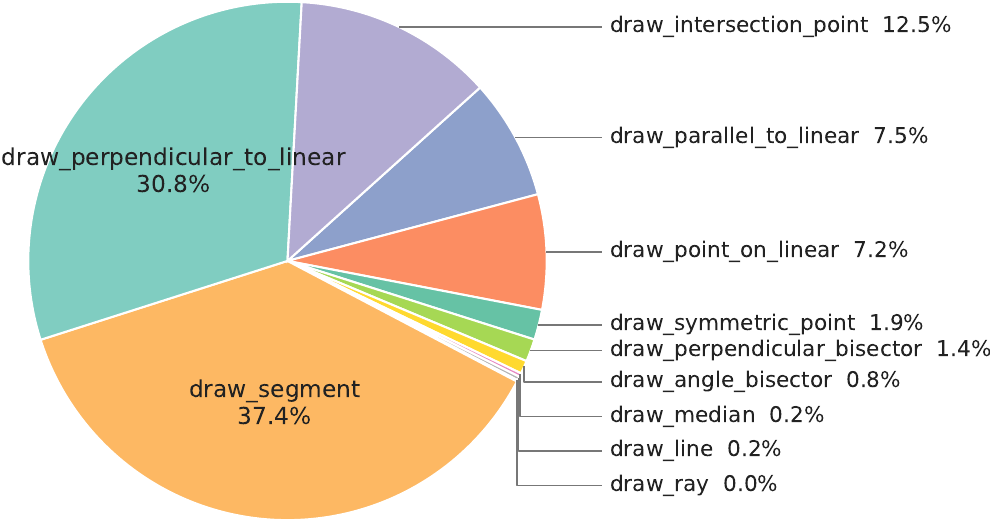}
\caption{Usage frequency of tools in the extended tool set across the training trajectories.}
\label{fig:tool_frequency}
\end{figure}

We initially considered an extended tool space covering a broader range of common geometric constructions. However, a larger action space increases prompt length and introduces additional ambiguity in tool selection during training. We therefore analyzed the tool usage distribution in the training trajectories, as shown in Fig.~\ref{fig:tool_frequency}, and retained the three most frequently used operations:
\texttt{draw\_segment},
\texttt{draw\_intersection\_point}, and
\texttt{draw\_perpendicular\_to\_linear}.

This design represents a trade-off between construction coverage and training stability rather than an inherent limitation of MetaReason. The framework can be extended to support additional constructions by incorporating corresponding tool definitions and updating the meta-information schema and renderer interfaces. The central contribution of MetaReason lies in the combination of structured geometric states, tool-based editing, and deterministic rendering, rather than in a fixed set of tools.

\section{Accuracy of MetaConverter}

We evaluate the robustness of MetaConverter through a human consistency study. We randomly sample 20\% of the examples from ExamGeo, GeoLaux-mini, and GeoQA. Three PhD-level annotators with mathematical backgrounds independently assess whether each re-rendered diagram preserves the geometric structure, point-line relations, and key annotations of the original diagram. The resulting strict diagram-level consistency is reported in Table~\ref{tab:metaconverter_consistency}.

\begin{table}[H]
\centering
\caption{Strict diagram-level consistency of MetaConverter on different benchmarks.}
\label{tab:metaconverter_consistency}
\small
\setlength{\tabcolsep}{4pt}
\begin{tabular}{lccc}
\toprule
Benchmark & ExamGeo & GeoLaux-mini & GeoQA \\
\midrule
Consistency & 52.5\% & 71.2\% & 82.0\% \\
\bottomrule
\end{tabular}
\end{table}

The lower consistency on ExamGeo is primarily attributable to its more challenging problems and more complex geometric diagrams. To reduce the impact of inaccurate meta-information on subsequent reasoning, MetaReason applies MetaJudge to filter inconsistent reconstructions before tool-augmented visual reasoning.

\section{Accuracy of MetaJudge}

We evaluate the reliability of MetaJudge through manual inspection by three PhD-level annotators. We report two complementary metrics. \textit{Acceptance precision} is the proportion of samples accepted by MetaJudge that are visually consistent with the original image. \textit{Inconsistent rejection rate} is the proportion of visually inconsistent reconstructions that are correctly rejected. The former measures the reliability of the samples retained for subsequent reasoning, while the latter directly measures MetaJudge's ability to detect reconstruction errors. The results are reported in Table~\ref{tab:metajudge_accuracy}.

\begin{table}[H]
\centering
\caption{Performance of MetaJudge on different benchmarks.}
\label{tab:metajudge_accuracy}
\small
\setlength{\tabcolsep}{4pt}
\begin{tabular}{lccc}
\toprule
Benchmark & ExamGeo & GeoLaux-mini & GeoQA \\
\midrule
Acceptance Precision & 98.0\% & 90.9\% & 94.5\% \\
Inconsistent Rejection & 98.5\% & 93.9\% & 94.0\% \\
\bottomrule
\end{tabular}
\end{table}

The high acceptance precision indicates that few inconsistent reconstructions are allowed to enter the tool-augmented reasoning stage. Among the inconsistent reconstructions that are mistakenly accepted, the errors mainly involve minor point or line displacements, redundant geometric objects, or inaccuracies in regions unrelated to the solution. Such errors do not necessarily alter the key geometric relations required for reasoning. The inconsistent rejection rate further reflects how effectively MetaJudge identifies and filters erroneous reconstructions.

\section{GeoQA Category Distribution}

The GeoQA test set contains 5,010 examples: 2,745 Angle, 323 Area, 1,873 Length, and 69 Other examples. The Other category therefore represents only 1.4\% of the test set. Its accuracy can change substantially with only a few different predictions, so comparisons on this subset should be interpreted cautiously.

\section{Case Study}
\makeatletter
\setlength{\@fptop}{0pt}
\makeatother
\begin{figure}[!htbp]
  \centering
  \includegraphics[width=\linewidth]{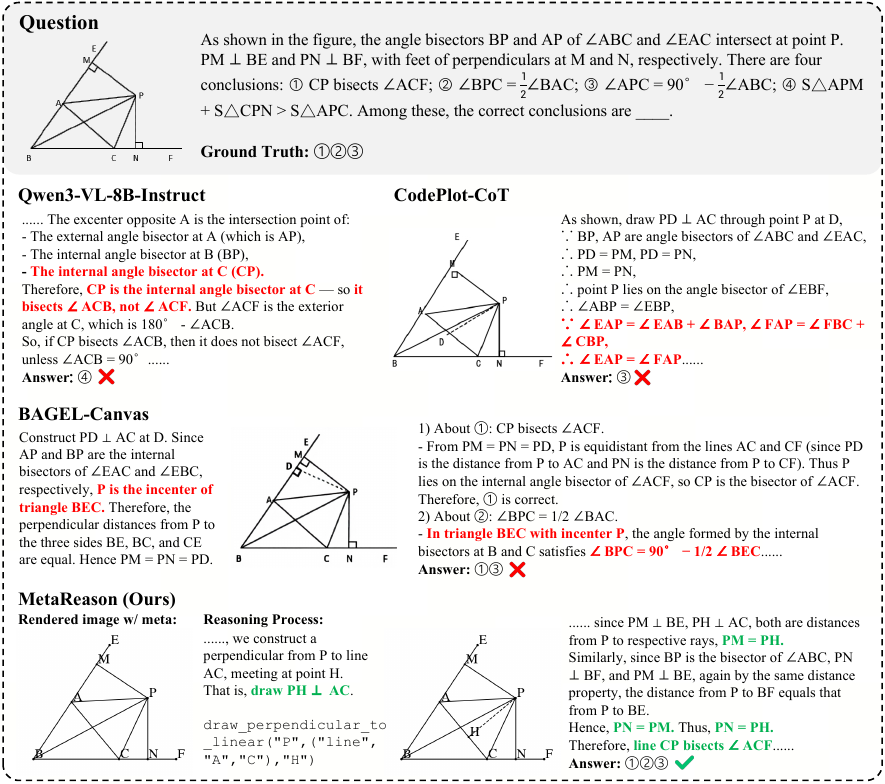}
  \caption{Case study of Geometric Reasoning on ExamGeo.}
  \label{fig:case_study_examgeo}
\end{figure}

\begin{figure}[!htbp]
  \centering
  \includegraphics[width=\linewidth]{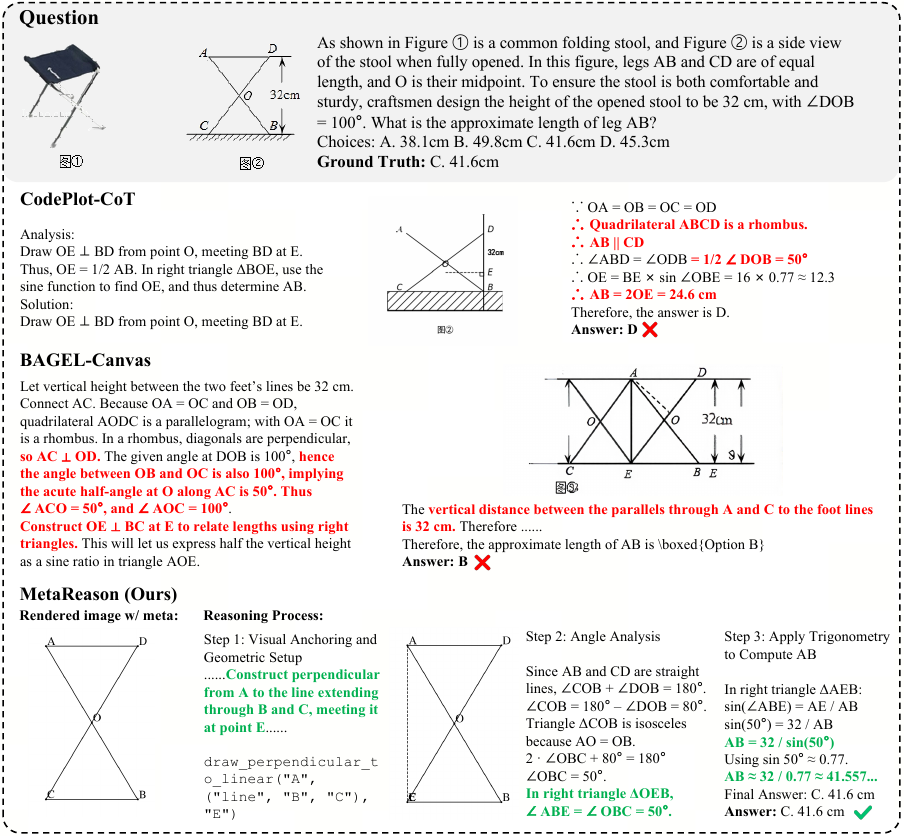}
  \caption{Case study of Geometric Reasoning on GeoQA}
  \label{fig:case_study_geoqa}
\end{figure}

\begin{figure}[!htbp]
  \centering
  \includegraphics[width=\linewidth]{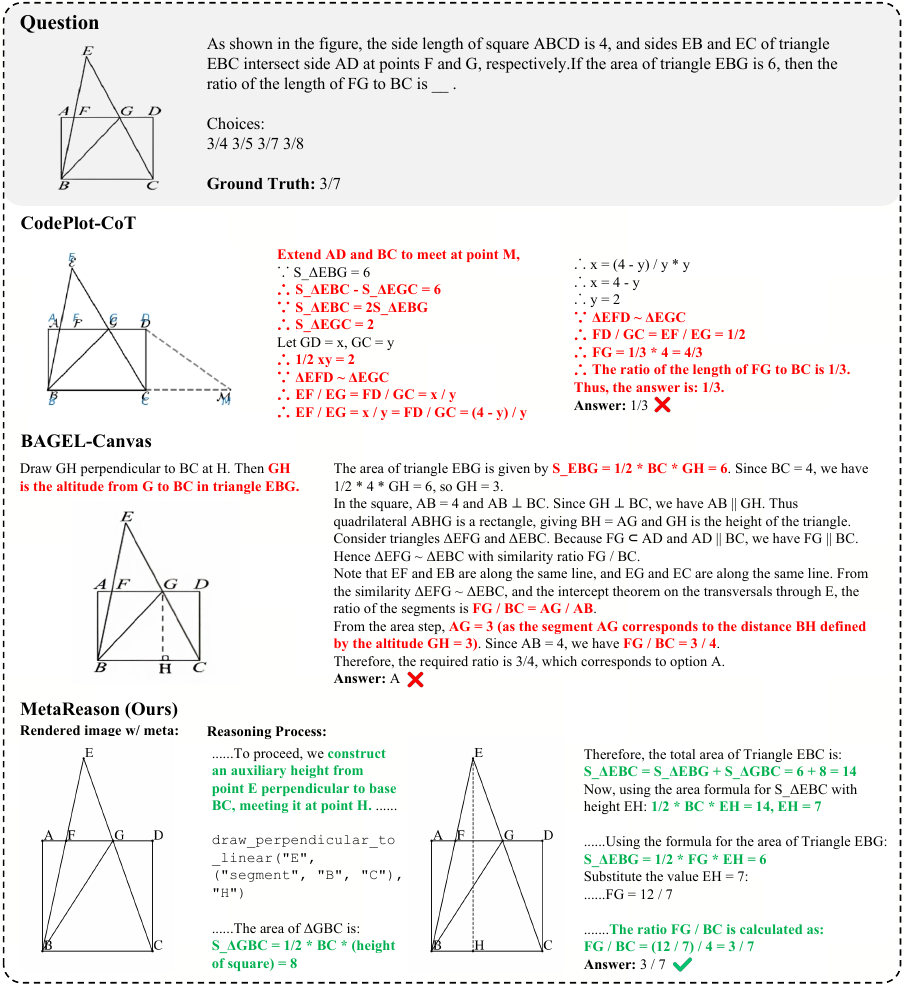}
  \caption{Case study of Geometric Reasoning on GeoLaux-mini. The aspect ratio of this figure is consistent with the original image provided in the GeoLaux-mini dataset. No scaling or post-processing has been applied, despite the visible vertical compression.}
  \label{fig:case_study_geolaux}
\end{figure}

\FloatBarrier
\makeatletter
\setlength{\@fptop}{0pt plus 1fil}
\makeatother

\section{Codes}
\subsection{Code for Rendering Meta-Information}
\begin{lstlisting}[
    language=Python,
    basicstyle=\ttfamily\small, % 附录可以用小号字，看得清楚
    breaklines=true,
    numbers=left,
    frame=tb,
    keepspaces=true, % 保留代码原有的缩进空格
    columns=flexible, % 核心：让空格宽度自然，不要死板的等宽对齐
    showstringspaces=false % 核心：关闭字符串中空格的可见符号
]
class _GeometryDrawer:
    def __init__(self, meta_data, min_size=512, max_size=1024, draw_points=True):
        self.meta_data = meta_data
        self.min_size = min_size
        self.max_size = max_size
        self.points = meta_data.get('points', {})
        self.shapes = meta_data.get('shapes', [])
        self.linewidth = max(min_size / 256, 1)
        self.fontsize = self.min_size / 16
        self.fontname = self._get_fontname()
        self.draw_points = draw_points
        self.fig, self.ax = None, None
        self._calculate_bounds()

    def _get_fontname(self):
        """Return a preferred Times-like font if available."""
        preferred_fonts = ['Times New Roman', 'Nimbus Roman', 'STIX', 'P052']
        available_fonts = set(f.name for f in font_manager.fontManager.ttflist)

        for font in preferred_fonts:
            if font in available_fonts:
                return font

        return 'serif'

    def _calculate_bounds(self):
        if not self.points:
            self.min_x = self.min_y = 0
            self.max_x = self.max_y = 10
        else:
            x_vals = [p[0] for p in self.points.values()]
            y_vals = [p[1] for p in self.points.values()]
            self.min_x, self.max_x = min(x_vals), max(x_vals)
            self.min_y, self.max_y = min(y_vals), max(y_vals)

            for shape in self.shapes:
                if shape.get('type') == 'circle':
                    center = shape.get('center')
                    radius = shape.get('radius', 1)
                    if center in self.points:
                        cx, cy = self.points[center]
                        self.min_x = min(self.min_x, cx - radius)
                        self.max_x = max(self.max_x, cx + radius)
                        self.min_y = min(self.min_y, cy - radius)
                        self.max_y = max(self.max_y, cy + radius)

        self.width = self.max_x - self.min_x
        self.height = self.max_y - self.min_y
        self.margin = max(self.width, self.height) * 0.1 or 1
        self.x_range = self.width + 2 * self.margin
        self.y_range = self.height + 2 * self.margin

    def _init_figure(self):
        aspect_ratio = self.x_range / self.y_range
        if self.x_range <= self.y_range:
            fig_width = self.min_size / 100
            fig_height = fig_width / aspect_ratio
        else:
            fig_height = self.min_size / 100
            fig_width = fig_height * aspect_ratio

        if self.max_size is not None and self.max_size > 0:
            if fig_width * 100 > self.max_size:
                fig_width = self.max_size / 100
            if fig_height * 100 > self.max_size:
                fig_height = self.max_size / 100

        self.fig, self.ax = plt.subplots(figsize=(fig_width, fig_height), dpi=100)
        self.ax.set_xlim(self.min_x - self.margin, self.max_x + self.margin)
        self.ax.set_ylim(self.min_y - self.margin, self.max_y + self.margin)
        self.ax.set_aspect('equal')
        self.ax.axis('off')

    def _draw_segment(self, shape, linestyle):
        p1, p2 = shape.get('points', [])
        if p1 in self.points and p2 in self.points:
            x1, y1 = self.points[p1]
            x2, y2 = self.points[p2]
            self.ax.plot([x1, x2], [y1, y2], linestyle=linestyle,
                         color='black', linewidth=self.linewidth)

    def _draw_line(self, shape, linestyle):
        p1, p2 = shape.get('points', [])
        if p1 in self.points and p2 in self.points:
            x1, y1 = self.points[p1]
            x2, y2 = self.points[p2]
            dx, dy = x2 - x1, y2 - y1
            boundary_points = []

            if abs(dx) > 1e-9:
                for x in self.ax.get_xlim():
                    y = y1 + (x - x1) * dy / dx
                    if self.ax.get_ylim()[0] <= y <= self.ax.get_ylim()[1]:
                        boundary_points.append((x, y))

            if abs(dy) > 1e-9:
                for y in self.ax.get_ylim():
                    x = x1 + (y - y1) * dx / dy
                    if self.ax.get_xlim()[0] <= x <= self.ax.get_xlim()[1]:
                        boundary_points.append((x, y))

            if len(boundary_points) >= 2:
                p1, p2 = boundary_points[:2]
                self.ax.plot([p1[0], p2[0]], [p1[1], p2[1]], linestyle=linestyle,
                             color='black', linewidth=self.linewidth)

    def _draw_circle(self, shape, linestyle):
        center = shape.get('center')
        radius = shape.get('radius', 1)
        if center in self.points:
            cx, cy = self.points[center]
            circle = patches.Circle((cx, cy), radius, fill=False, edgecolor='black',
                                    linestyle=linestyle, linewidth=self.linewidth)
            self.ax.add_patch(circle)

    def _draw_shapes(self):
        style_map = {'solid': '-', 'dashed': '--'}
        for shape in self.shapes:
            shape_type = shape.get('type')
            style = shape.get('style', 'solid')
            linestyle = style_map.get(style, '-')

            if shape_type == 'segment':
                self._draw_segment(shape, linestyle)
            elif shape_type == 'line':
                self._draw_line(shape, linestyle)
            elif shape_type == 'circle':
                self._draw_circle(shape, linestyle)

    def _draw_points(self):
        for name, (x, y) in self.points.items():
            if name != "":
                self.ax.scatter(x, y, s=20, c='black', zorder=5)
                self.ax.annotate(name, (x, y), xytext=(3, 3), textcoords='offset points',
                                fontsize=self.fontsize,
                                fontname=self.fontname)
            else:
                pass

    def _draw(self):
        with _PLOT_LOCK:
            try:
                self._init_figure()
                self._draw_shapes()
                if self.draw_points:
                    self._draw_points()
                plt.tight_layout()
                
                buf = io.BytesIO()
                self.fig.savefig(buf, format='png', dpi=100, bbox_inches='tight')
                buf.seek(0)
                pil_image = Image.open(buf).convert('RGB')
                return pil_image
            finally:
                if self.fig:
                    plt.close(self.fig)

def draw_geometry(meta_data, min_size=512, max_size=1024, draw_points=True, return_none_with_error=True):
    """
    Draw geometry from metadata.

    Args:
        meta_data (dict): Points and shapes.
        min_size (int): Minimum image side length.
        max_size (int): Maximum image side length in pixels.
        draw_points (bool): Whether to draw point markers and labels.
        return_none_with_error (bool): Return None on error if True; otherwise raise.

    Returns:
        PIL.Image.Image | None: Rendered image or None on failure.
    """
    try:
        drawer = _GeometryDrawer(meta_data, min_size=min_size, max_size=max_size, draw_points=draw_points)
        return drawer._draw()
    except Exception as e:
        if not return_none_with_error:
            raise e
        else:
            return None
\end{lstlisting}

\subsection{Implementation Detail of Toolbox}
\begin{lstlisting}[
    language=Python,
    basicstyle=\ttfamily\small,
    breaklines=true,
    numbers=left,
    frame=tb,
    keepspaces=true, % 保留代码原有的缩进空格
    columns=flexible, % 核心：让空格宽度自然，不要死板的等宽对齐
    showstringspaces=false % 核心：关闭字符串中空格的可见符号
]
class Geometry:
    def __init__(self, meta_data=None):
        self.meta_data = meta_data

    def _get_coord(self, p: str):
        """Return the coordinate of point p."""
        return self.meta_data["points"][p]

    def _add_point(self, coord, name: str):
        """Register a named point in meta_data."""
        self.meta_data["points"][name] = coord
        return name

    def _line_intersection(self, A, B, C, D):
        """Return the intersection of lines AB and CD."""
        ax, ay = self._get_coord(A)
        bx, by = self._get_coord(B)
        cx, cy = self._get_coord(C)
        dx, dy = self._get_coord(D)

        denom = (bx-ax)*(dy-cy) - (by-ay)*(dx-cx)
        if abs(denom) < 1e-9:
            raise ValueError("Lines are parallel or coincident, no unique intersection point.")

        t = ((cx-ax)*(dy-cy) - (cy-ay)*(dx-cx)) / denom
        return ax + t*(bx-ax), ay + t*(by-ay)

    def _in_segment(self, Pxy, A, B):
        """Check whether point Pxy lies on segment AB."""
        px, py = Pxy
        ax, ay = self._get_coord(A)
        bx, by = self._get_coord(B)

        if abs((px-ax)*(by-ay) - (py-ay)*(bx-ax)) > 1e-9:
            return False
        return (px-ax)*(px-bx) + (py-ay)*(py-by) <= 1e-9

    def connect_points(self, linear: tuple):
        """Append a line/segment shape from two existing points."""
        type, point1, point2 = linear
        assert type in ("line", "segment"), "Unsupported linear type: {}".format(type)
        assert point1 in self.meta_data["points"], "Point {} not found in meta_data. Available points {}".format(point1, list(self.meta_data["points"].keys()))
        assert point2 in self.meta_data["points"], "Point {} not found in meta_data. Available points {}".format(point2, list(self.meta_data["points"].keys()))
        self.meta_data["shapes"].append({
            "type": type,
            "points": [point1, point2],
            "style": "dashed"
        })

    def draw_intersection_point(self, linear1: tuple, linear2: tuple, name: str):
        """Draw an intersection point of two linear elements."""
        if name in self.meta_data["points"]:
            raise ValueError("Point name '{}' already exists in meta_data. Please choose a different name.".format(name))

        t1, A1, B1 = linear1
        t2, A2, B2 = linear2
        P = self._line_intersection(A1, B1, A2, B2)
        ok1 = (t1 == "line") or (t1 == "segment" and self._in_segment(P, A1, B1))
        ok2 = (t2 == "line") or (t2 == "segment" and self._in_segment(P, A2, B2))
        assert ok1 and ok2, "No intersection point found for the given linear elements, linear1: {}, linear2: {}".format(linear1, linear2)
        self._add_point([P[0], P[1]], name=name)

    def draw_segment(self, point1: str, point2: str):
        self.connect_points(("segment", point1, point2))

    def draw_perpendicular_to_linear(self, point: str, linear: tuple, name: str):
        """Draw a perpendicular from point to a linear element and add the foot point."""
        if name in self.meta_data["points"]:
            raise ValueError("Point name '{}' already exists in meta_data. Please choose a different name.".format(name))

        t, A, B = linear
        px, py = self._get_coord(point)
        ax, ay = self._get_coord(A)
        bx, by = self._get_coord(B)
        dx, dy = bx - ax, by - ay
        assert dx != 0 or dy != 0, "target linear element cannot be a degenerate point, A: {}, B: {}".format(A, B)
        apx, apy = px - ax, py - ay
        L2 = dx*dx + dy*dy
        t0 = (apx * dx + apy * dy) / L2
        hx = ax + t0 * dx
        hy = ay + t0 * dy
        H = self._add_point([hx, hy], name=name)
        self.connect_points(("segment", point, H))
\end{lstlisting}

\section{Prompts}
\subsection{Prompt for MetaConverter}

\begin{tcolorbox}[
    breakable,
    title=\textbf{Prompt for MetaConverter},
]
\textbf{\textit{SYSTEM PROMPT:}} \\
You are an expert in geometric figure recognition, skilled at extracting configuration information from geometry problem statements and, without solving the problem, converting the figure in the problem image into a JSON-formatted geometric meta description. Please strictly follow the rules below to complete the task: \\
\vspace{0.3em}

\textbf{\textit{USER PROMPT:}} \\
\textbf{[Task Objective]} \\
Based on the problem statement text and the problem image, extract all key points and geometric relationships, and output JSON meta information representing the figure structure. There is no need to solve numerical values or provide proofs. The extracted meta information should remain as consistent with the original image as possible. \\
\vspace{0.3em}

\textbf{[Reasoning Requirements]} \\
1. Analyze the textual information in the problem statement step by step and, together with the image, determine the coordinates of each point and the geometric objects to be drawn.\\
2. Assign reasonable coordinates according to the image layout, keeping geometric relationships correct:\\
\quad - Prefer integers or simple fractions during reasoning; when outputting JSON, convert fractions to floating-point numbers (keep up to 3 decimal places).\\
\quad - For points and shapes whose positions are not explicitly specified in the problem statement, preferentially choose ordinary positions that roughly match the original layout so the overall figure resembles the original. Avoid placing such points or shapes in obviously special positions (for example exact midpoints, perpendicular feet, circle centers, or alignments that force strict parallelism/perpendicularity), unless the problem statement explicitly requires it.\\
\quad - For geometric relationships that cannot be directly deduced or are not obtainable by simple calculations, you may choose not to use them to draw the figure. Coordinates for the involved points may be determined based on the image so that the reconstructed figure resembles the original; exact consistency is not required.\\
\quad - The reconstructed figure should match the original in structure and layout as closely as possible. Do not draw lines or shapes that are not present in the original image; conversely, every visible point, line, or other shape in the original image must be reflected in the reconstructed meta information.\\
3. Coordinate convention: with respect to the image, right is the positive x direction and up is the positive y direction.\\
4. Point naming must be combinations of letters, digits, apostrophes, and underscores (for example $A$, $B^\prime$, $O_1$), and names should be kept concise. Avoid overly complex names such as $A^\prime$, $B_{30}$ (multiple numbers), $V_{1,1}$, $P^\prime_0$, etc.\\
5. The final output must be valid JSON, and no explanatory text is allowed inside the JSON. \\
\vspace{0.3em}

\textbf{[JSON Output Structure]} \\
The output must follow the following format (where ``points'' maps point names to [x, y] coordinates, and "shapes" lists geometric objects; see the geometry types below for details): \\
\vspace{0.3em}

\begin{verbatim}
```json
{
  "points": {
    "<pointName1>": [<x1>, <y1>],
    ...
  },
  "shapes": [
    <shape1>,
    ...
  ]
}
\end{verbatim}
\vspace{0.3em}

Be sure to include the \verb|```|\texttt{json} and \verb|```| markers to ensure the output is a valid JSON code block. Otherwise the output cannot be parsed. \\
\vspace{0.3em}

\textbf{[Supported Geometric Figure Types]} \\
Notes:\\
- All geometric objects in JSON are represented as objects and must include a ``type'' field specifying the object type.\\
- All fields that represent points must use point-name strings (for example ``$A$'', ``$B_1$'', ``$C^\prime$'').\\
- All points lists must be arrays of strings.\\
- The ``style'' field accepts only two string values: ``solid'' or ``dashed''. \\
\vspace{0.3em}

1) segment\\
- Description: a line segment connecting two endpoints.\\
- Required fields:\\
\quad - ``type'': ``segment''\\
\quad - ``points'': [``startPointName'', ``endPointName'']\\
\quad - ``style'': ``solid'' | ``dashed'' \\
\vspace{0.3em}

2) line\\
- Description: an infinite straight line determined by two points.\\
- Required fields:\\
\quad - ``type'': ``line''\\
\quad - ``points'': [``pointName1'', ``pointName2'']\\
\quad - ``style'': ``solid'' | ``dashed'' \\
\vspace{0.3em}

3) circle\\
- Description: a circle defined by its center and radius.\\
- Required fields:\\
\quad - ``type'': ``circle''\\
\quad - ``center'': ``centerPointName''\\
\quad - ``radius'': number\\
\quad - ``style'': "solid" | "dashed" \\
\vspace{0.3em}

Reminders:\\
- All numeric values must be numeric; string representations of numbers are not accepted.\\
- Ensure all points referenced in the ``shapes'' list are defined in the ``points'' dictionary.\\
- Ensure that all necessary geometric figures are included in the ``shapes'' list; avoid defining only points while forgetting to define the geometric objects that connect them.\\
- Please strictly adhere to the original image content: do not introduce any lines or shapes that do not exist in the original image; likewise, every line or shape visible in the original image must be represented. \\
\vspace{0.3em}

\textbf{[Your Task]} \\
Following the requirements above, and based on the following problem statement text and problem image, analyze the statement step by step, infer coordinates for each vertex of the geometric figure, and then output the resulting meta information in valid JSON format. 
\end{tcolorbox}

\subsection{Prompt for MetaJudge}
\begin{tcolorbox}[
    breakable,
    title=\textbf{Prompt for MetaJudge},
]
\textbf{\textit{SYSTEM PROMPT:}} \\
You are a professional geometry expert responsible for evaluating whether a [Reconstructed Image] of a geometry problem is reasonable with respect to the [Question Image] and the [Question Text]. Please \textbf{do not solve the problem}; simply determine whether the reconstructed image is rational. Follow the evaluation steps strictly as outlined below: \\
\vspace{0.3em}

\textbf{\textit{USER PROMPT:}} \\
\textbf{[Evaluation Steps]} \\
Step 1: Topological Comparison (Ignore Question Text) \\
Directly compare the geometric structure (shape types, point naming, connection relationships, relative positions) of the Question Image and the Reconstructed Image. \\
- Judge as Rational (\verb|\boxed{False}|): \\
\quad - The topological structure is identical. \\
\quad - The Reconstructed Image is simply a rotation, translation, reflection, or scaling (congruent or similarity transformation) of the Question Image while preserving all of the position relationships between points and shapes. \\
- Judge as Irrational (\verb|\boxed{True}|): \\
\quad - Inconsistent Points: The Reconstructed Image misses points from the Question Image; its point labels are inconsistent with the Question Image, or the positions of labeled points do not match the Question Image (e.g., intersections between segments or circles are marked on non-intersection points; points that should be on a shape like segment or circle significantly deviate from that shape or are marked on a different shape). \\
\quad - Isolated points: The Reconstructed Image has any labeled point that is an isolated point, meaning it does not lie on any line or circle. \\
\quad - Inconsistent Connections: The Reconstructed Image has connections not present in the Question Image, or is missing connections that exist in the Question Image (excluding auxiliary markings like angles or perpendicularity). \\
\vspace{0.3em}

Step 2: Question Constraint Verification (When structural differences exist) \\
If the Reconstructed Image and the Question Image do not satisfy the ``structural consistency'' mentioned above, verification must be performed in conjunction with the [Question Text]: \\
- Judge as Rational (\verb|\boxed{False}|): \\
\quad - The Reconstructed Image fully satisfies the explicit geometric constraints given in the question (e.g., a point lying on a specific segment, parallelism, perpendicularity, proportional relationships, etc.). \\
\quad - Tolerance for Specialization: If the question does not specify a particular position for a point/line, and the Reconstructed Image places it in a special position (e.g., midpoint, angle bisector), it is considered rational as long as it does not violate known conditions. \\
\quad - Visual Tolerance: Maintain leniency toward qualitative errors such as segments' lengths and angles' degrees. Unless there is an obvious qualitative violation (e.g., an acute angle drawn as obtuse, equal segments having significantly different lengths, or a midpoint being clearly offset), it is considered satisfied. \\
\quad - Extra Point Tolerance: If the Reconstructed Image labels points not mentioned in the question, it is considered rational as long as it does not violate known conditions. \\
- Judge as Irrational (\verb|\boxed{True}|): \\
\quad - Violation of Constraints: The Reconstructed Image violates any explicit geometric relationship described in the question. \\
\quad - Information Missing: Differences exist between the two images, and the question text provides no geometric description to verify them. \\
\vspace{0.3em}

\textbf{[Output Requirements]} \\
1. Analysis Process: Briefly explain the basis of your judgment. \\
2. Final Conclusion: Output the judgment result in the very last line. \\
\quad - \verb|\boxed{True}|: Represents irrational (to be filtered). \\
\quad - \verb|\boxed{False}|: Represents rational (to be retained). \\
\vspace{0.3em}

Note: Strictly adhere to the LaTeX formatting requirements. Do not change the case, add spaces, or include extra decorations (e.g., do not output \verb|\boxed{True}| or \verb|\boxed{False}|). \\
\vspace{0.3em}

\textbf{[Question Text]} \\
$<$QUESTION\_TEXT$>$ \\
\vspace{0.3em}

\textbf{[Question Image]}\\
$<$QUESTION\_IMAGE$>$ \\
\vspace{0.3em}

\textbf{[Reconstructed Image]}\\
$<$RECONSTRUCTED\_IMAGE$>$ \\
\vspace{0.3em}

Note again: please DO NOT solve the problem; simply determine whether the reconstructed image is rational. Be a strict critic, not a helper. If you suspect any part of it is irrational, please decisively judge it as irrational and output \verb|\boxed{True}|. Now, please begin your evaluation:
\end{tcolorbox}

\subsection{Prompt for MetaReasoner to Implement Multimodal Reasoning}

\begin{tcolorbox}[
    breakable,
    title=\textbf{Prompt for MetaReasoner to Implement Multimodal Reasoning},
]
\textbf{\textit{SYSTEM PROMPT:}} \\
You are an expert in plane geometry problem solving. Your task is to perform rigorous and coherent geometric reasoning based on the problem description and the geometric image, and to provide a complete solution process and final conclusion. \\
You can use a ``Geometry Sketchpad'' to assist your reasoning. The sketchpad is operated through a function interface named \texttt{geometry}, which can construct auxiliary lines, intersection points, and perpendicular lines on the figure, but cannot directly read or modify the original image. \\
\vspace{0.3em}

\textbf{\textit{USER PROMPT:}} \\
\textbf{[General Principles]} \\
- Use the Geometry Sketchpad only when it is truly necessary to construct auxiliary lines; do not call tools without purpose. \\
- Once you call a tool, all subsequent reasoning must continue based on the updated figure. \\
\vspace{0.3em}

\textbf{[Problem-Solving Workflow]} \\
1. $<$think$>$ Reasoning \\
\quad - Analyze the problem conditions and derive geometric relationships. \\
\quad - Decide whether auxiliary constructions are needed and whether the Geometry Sketchpad should be used. \\
\vspace{0.3em}

2. $<$tool\_call$>$ Drawing (if needed) \\
\quad - Use it only when auxiliary constructions or figure updates are needed. \\
\quad - The $<$tool\_call$>$ block must contain only Python tool-call code for the Geometry Sketchpad and nothing else. \\
\vspace{0.3em}

3. $<$answer$>$ Conclusion \\
\quad - Provide the final answer or proof conclusion inside $<$answer$>$$\backslash$boxed\{...\}$<$/answer$>$. \\
\vspace{0.3em}

\textbf{[Available Geometry Sketchpad Functions]} \\
\texttt{geometry.draw\_segment(start\_point\_name: str, end\_point\_name: str)} \\
- Function: Connect two points to draw a segment. \\
\vspace{0.3em}

\texttt{geometry.draw\_intersection\_point(linear1: linear, linear2: linear, new\_point\_name: str)} \\
- Function: Compute the intersection of two lines/segments and name it \texttt{new\_point\_name}. \\
- Parameters: \texttt{linear} is defined as (``line''/``segment'', \texttt{point\_A}, \texttt{point\_B}). \\
- Note: This function only adds a point. If you need a segment from an endpoint to the intersection, call \texttt{draw\_segment} additionally. \\
\vspace{0.3em}

\texttt{geometry.draw\_perpendicular\_to\_linear(start\_point: str, target\_linear: linear, foot\_point\_name: str)} \\
- Function: Draw a perpendicular from \texttt{start\_point} to \texttt{target\_linear} and name the foot point as \texttt{foot\_point\_name}. \\
\vspace{0.3em}

\textbf{[Drawing Rules]} \\
- Newly constructed points must not overwrite existing points. \\
- Once a tool call modifies the figure, subsequent reasoning should be based on the updated figure (including new auxiliary lines). \\
\vspace{0.3em}

Please start solving the problem.
\end{tcolorbox}

\subsection{Prompt for MetaReasoner to Implement Text-only Reasoning}
\begin{tcolorbox}[
    breakable,
    title=\textbf{Prompt for MetaReasoner to Implement Text-only Reasoning},
]
\textbf{\textit{SYSTEM PROMPT:}} \\
You are a senior expert in plane geometry proficient in problem-solving. Your task is to perform rigorous and coherent geometric reasoning based on the given geometric image and problem description, providing detailed solution steps and conclusions. You should simulate human problem-solving thinking: deeply analyze known conditions, prudently construct auxiliary lines, keenly capture geometric correlations, and finally provide an accurate answer or a rigorous proof. \\
\vspace{0.3em}

\textbf{\textit{USER PROMPT:}} \\
After the reasoning is finished, provide the final answer or proof conclusion in the \texttt{$<$answer$>$$\backslash$boxed\{...\}$<$/answer$>$} tags. \\
\vspace{0.3em}

Now, please start solving the following problem according to the above specifications.
\end{tcolorbox}


\end{document}